\documentclass[a4paper,conference]{IEEEtran}

\usepackage{cite}

\ifCLASSINFOpdf
   \usepackage[pdftex]{graphicx}
   \usepackage{subfig}
   \usepackage{bigstrut}
\else
\fi

\usepackage{amsmath}
\usepackage{amssymb}
\usepackage{xcolor}
\usepackage{wrapfig}
\usepackage{multicol}
\usepackage{enumitem}
\usepackage{todonotes}
\usepackage[normalem]{ulem}
\usepackage{booktabs}
\usepackage[font=small]{caption}

\usepackage{contour}
\newcommand{\myparagraph}[1]{\smallskip\noindent\textbf{#1:}}

\usepackage{soul}
\usepackage{booktabs}
\usepackage{xcolor,colortbl}

\begin{document}
%

\title{Facial Tic Detection in Untrimmed Videos of Tourette Syndrome Patients}

\author{\IEEEauthorblockN{Yutao Tang\IEEEauthorrefmark{1},
Benjamín B\'ejar\IEEEauthorrefmark{2},
Joey K.-Y. Essoe\IEEEauthorrefmark{3},
Joseph F. McGuire\IEEEauthorrefmark{3} and
René Vidal\IEEEauthorrefmark{1}
}

\IEEEauthorblockA{\IEEEauthorrefmark{1} Mathematical Institute for Data Science and Department of Biomedical Engineering, Johns Hopkins University, USA}
\IEEEauthorblockA{\IEEEauthorrefmark{2} Swiss Data Science Center, Paul Scherrer Institut, Switzerland}
\IEEEauthorblockA{\IEEEauthorrefmark{3} Department of Psychiatry and Behavioral Sciences, Johns Hopkins University School of Medicine, USA}
\IEEEauthorblockA{\{ytang67, rvidal\}@jhu.edu, benjamin.bejar@psi.ch, \{essoe, jfmcguire\}@jhmi.edu.
}

}

\maketitle

\begin{abstract}
Tourette Syndrome (TS) is a behavioral disorder that onsets in childhood and is characterized by the expression of involuntary movements and sounds commonly referred to as tics. Behavioral therapy is the first-line treatment for patients with TS, and it helps patients raise awareness about tic occurrence as well as develop tic inhibition strategies. However, the limited availability of therapists and the difficulties for in-home follow up work limits its effectiveness. An automatic tic detection system that is easy to deploy could alleviate the difficulties of home-therapy by providing feedback to the patients while exercising tic awareness. In this work, we propose a novel architecture (T-Net) for automatic tic detection and classification from untrimmed videos. T-Net combines temporal detection and segmentation and operates on features that are interpretable to a clinician. We compare T-Net to several state-of-the-art systems working on deep features extracted from the raw videos and T-Net achieves comparable performance in terms of average precision while relying on interpretable features needed in clinical practice.

\end{abstract}


%
\IEEEpeerreviewmaketitle

\section{Introduction}
Tourette Syndrome (TS) is a neuropsychiatric condition characterized by involuntary movements (motor tics) and vocalizations (vocal tics) \cite{Bloch:2009},  which can cause pain, injury, stress, or serious social impairment \cite{Conelea:2011, Conelea:2013}. Tics usually start in the facial areas with eye movements being the most common tics \cite{Mcguire:2013}. Although pharmacotherapies are available, behavioral therapy is universally recommended as a first-line treatment because of its efficacy and low adverse effects 
\cite{Essoe:2019, Weisman:2013, Pringsheim:2019}.

In behavioral therapy, therapeutic improvement depends on patients learning behavioral skills to reduce tic expression. The two central skills are the development of tic awareness and the implementation of behavioral strategies that inhibit tic expression contingent upon such awareness. To master these skills, patients are assigned “homework” practices between therapy sessions. However, as many patients with TS have limited tic awareness, a support partner (e.g., a parent) is required during homework practices to help patients identify tic occurrences. Unfortunately, limitations in the availability and tic identification accuracy of the support partners are often barriers to homework practices. Since the success of homework practices predicts therapeutic improvements and treatment response \cite{Essoe:2021} to behavioral therapy, tools that address these barriers hold considerable promise.
A possible solution to this problem would be to have an easy-to-deploy automatic system that is able to detect and inform patients about the occurrences of tics. Such a system would lower the barriers for homework practice and bear the promise of improving therapy outcomes. A few studies considered the problem of automatic tic detection \cite{Shute:2016, Barua:2021, Wu:2021}. 
\cite{Shute:2016} used support vector machines to classify tics based on segmented temporal windows of deep brain simulation signals. 
\cite{Barua:2021} used LeNet-5 to classify tics based on segmented temporal windows of wireless channel information signals. 
\cite{Wu:2021} used LSTM to classify tics based on well-trimmed temporal windows of deep features extracted from the videos of patients using a convolutional neural network. 

Despite the outstanding contributions of these studies, they have two main drawbacks. First, most studies rely on data that requires a complex device setup, which is hard to implement in common clinical settings or at the patients' home. Second, all prior studies performed tic classification on well-localized temporal windows, which ignores long-range temporal relevancy and limits the accuracy of the identification of tic occurrences. To overcome these shortcomings and promote the accessibility and accuracy of the tool, we propose to use videos recorded by a smartphone and perform temporal tic detection in untrimmed videos. Besides, as facial motor tics are very common among TS patients, we focus our attention on developing a tool for facial tic detection.

Facial tic detection from videos is a particular instance of activity detection. Activity detection is a significant and challenging task in the field of video analysis and has practical applications in video recommendation, smart surveillance, robotics, and beyond. Benefiting from the increased computational power and the vast success of deep learning, as well as the availability of large-scale public datasets with thousands of annotated videos, many successful temporal action detection methods \cite{Zeng:2019,Shou:2016,Lin:2019,Lin:2018,liu:2019mgg,lin2020dbg,Lin:2017ssad,Zhao:2017ssn,Zhang:2018s3d,Qing:2021tcanet,Wang:2021daotad} have been proposed. For example, \cite{Lin:2019} developed a boundary-sensitive approach, Boundary-Matching Network (BMN), for obtaining candidate action proposals. \cite{Qing:2021tcanet} proposed Temporal Context Aggregation Network (TCANet) that uses both local and global temporal context to progressively obtain accurate action boundaries and reliable probability predictions. \cite{Wang:2021daotad} combined many image-level data augmentations to design a data augmentation one-stage temporal action detector (DaoTAD).

However, tic detection poses several challenges that limit the use of state-of-the-art methods.
First, due to patient privacy and confidentiality, publicly available videos are infrequent and never annotated, while most deep learning methods require large datasets for training. Second, tics are extremely variable in duration and heterogeneous in expression across and within patients despite sharing the same category. For example, an ``eye blink" tic may vary in duration, repetition, and forcefulness across patients, or even vary week-to-week within the same person.
Third, the lack of interpretability presents another barrier in applying current deep learning techniques to tic detection as patients and clinicians may not be able to interpret results in the context of behavioral therapy.

\begin{figure*}
\centering
\setlength{\belowcaptionskip}{-0.3cm} 
\includegraphics[trim=0cm 0cm 0cm 0cm, clip,width=0.8\textwidth]{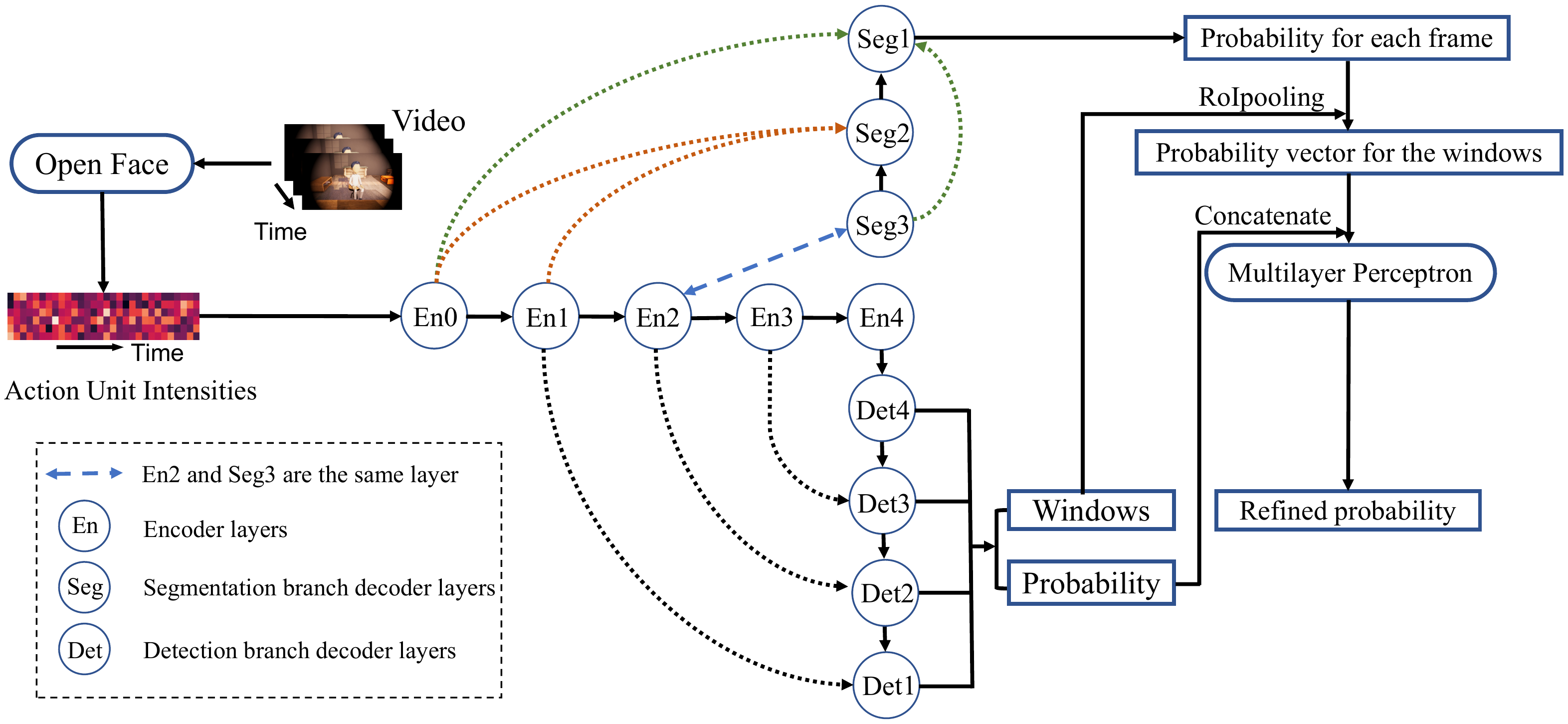}
\caption{Structure of T-Net (shared). ``En0" represents pre-residual convolutional layer, and ``En1" through ``En4" denote the four residual stages of the temporal ResNet-18. \textbf{Segmentation branch:} ``Seg1" through ``Seg3" represent the T-UNet3+ decoder layers. \textbf{Detection branch:} ``Det1" through ``Det4" represent the T-RetinaNet decoder layers (T-FPN). \textbf{Score Fusion Head:} 1D RoIPooling was performed on the output of ``Seg1" to create probability vectors, guided by the windows predicted from the detection branch. After that, these probability vectors were concatenated with the probability predictions from the detection branch. The concatenations were then processed by an MLP, generating a refined probability for each window.}
\label{fig:T-Net}
\end{figure*}

In this paper, we address the problem of tic detection in untrimmed videos obtained from a smartphone. Specifically, we first establish baseline systems based on BMN \cite{Lin:2019}, DaoTAD \cite{Wang:2021daotad} and TCANet \cite{Qing:2021tcanet} using deep features extracted from the raw videos. Then, to make our system interpretable to potential users, we use facial Action Unit (AU) intensities as features \cite{Ekman:1978AU}. AUs are localized facial movements, such as ``upper lip raiser" and ``nose wrinkler", representing the fundamental actions of individual facial muscles or muscle groups. 
AU intensities are clinically meaningful features as clinicians identify facial motor tics through the forcefulness, abruptness, and location of patient movements. 
However, directly extending the baseline systems to use AU intensities as input features degrade the performance. 
To reduce this drop of performance, we propose a novel Tic detection network (T-Net) that uses interpretable AU intensities as input features and achieves comparable performance to BMN and better performance than DaoTAD and TCANet, using deep features. As shown in Fig. \ref{fig:T-Net}, the proposed T-Net mainly consists of two branches. The \emph{temporal segmentation branch} is an encoder-decoder architecture that takes a sequence of AU intensities as input and predicts a tic probability at each frame. Hence, some post-processing is needed to convert segmentation probabilities into detection probabilities. The \emph{temporal detection branch} directly produces temporal windows with associated detection probabilities via an encoder-decoder architecture. However, the detection probabilities tend to be noisier than the segmentation probabilities. T-Net obtains the best of both worlds by fusing both segmentation and detection probabilities, while at the same time mitigating the challenges of applying state-of-the-art action detection methods to tic detection. Specifically, the need for large training sets is mitigated by using a compact architecture with a small number of layers as well as various data augmentation techniques.
Then, T-Net obtains precise boundaries for tics of variable duration by using k-means clustering to get more suitable anchor sizes and applying an additional regression loss, the efficient intersection over union (EIoU) loss \cite{Zhang:2021}, to penalize imprecise boundaries. Finally, T-Net produces interpretable detections by using AU intensities as features.




In summary, our contributions are three-fold:
\begin{enumerate}[leftmargin=*]
\item We study a novel application of temporal action detection, detection of facial tics in untrimmed videos, which requires extending extant methods to the small training data regime.

\item We propose a novel one-stage temporal detector, T-Net, that uses temporal segmentation probabilities to refine detection probabilities for temporal tic detection.

\item We demonstrate that using clinically meaningful AU intensities as input features maintains good performance, in addition to gaining
interpretability of the detection results.

\end{enumerate}

\section{Related work}
Our proposed T-Net architecture for tic detection integrates both temporal detection and temporal segmentation approaches, which we briefly review in this section.


%
\subsection{Temporal Action Detection}
The goal of temporal action detection is to find instances of an action in a video. Most temporal action detection methods draw inspiration from the object detection literature, where the goal is to find instances of an object in a 2D image. Object detectors can be classified into two main categories depending on the way they operate. On the one hand, two-stage detectors such as \cite{Girshick:2015, Ren:2015, He:2017maskrcnn, Cai:2018} follow the propose-then-classify scheme in which a number of candidate bounding boxes are first generated through a proposal generation network and subsequently classified. On the other hand, one-stage detectors like \cite{Lin:2017RetinaNet, Liu:2016, Redmon:2017, Tan:2020} bypass the proposal generation stage and directly detect objects. While two-stage detectors are usually more accurate, one-stage detectors are generally more efficient. 

Temporal action detection methods follow a similar approach to object detection methods, except that 2D bounding boxes are replaced by temporal windows. Specifically, two-stage methods, such as \cite{Shou:2016,Zeng:2019,Lin:2019, Lin:2018, Gao:2017turn, Gao:2018, liu:2019mgg, lin2020dbg}, learn a class-agnostic temporal action proposal generator and the generated proposals are then fed to an action classifier to produce the final detection results. 
Meanwhile, one-stage methods, such as
\cite{Lin:2017ssad, Zhao:2017ssn, Zhang:2018s3d, Xu:2020, Liu:2020pbr}, 
jointly learn the action proposal and classification networks, thus they are more efficient. 
Since our goal is to develop a compact model that can be trained from limited data, the detection branch of our T-Net uses a one-stage temporal action detector adapted from RetinaNet \cite{Lin:2017RetinaNet}.




\subsection{Temporal Action Segmentation}
The goal of temporal action segmentation is to predict an action label for each frame in a video. This requires capturing short- and long-range dependencies between frames. Popular models to capture such temporal dependencies include hidden Markov models \cite{Tao:2012,Kuehne:2016,Kuehne:2018}, conditional random fields \cite{Lea:2016learning, Mavroudi:2018}, recurrent neural networks \cite{Singh:2016}, and temporal convolutional networks \cite{Lea:2016,Lea:2017,Lei:2018}. However, the computational complexity of such approaches during training and inference limits their applicability to short-range dependencies, which can result in oversegmentation.
Multi-stage temporal convolutional networks \cite{Farha:2019,Li:2020,Wang:2020} alleviate this issue by using multiple stages of refinement. However, such networks have a very large number of parameters and thus they cannot achieve good performance with limited data.
Temporal semantic segmentation networks \cite{Wang:2019, Luktuke:2020} alleviate this problem by using multi-scale encoder-decoder architectures, which can also capture long-range interactions, and thus have the potential to produce more accurate estimate of the probabilities of a tic at the frame level. Therefore, in this work we use a temporal semantic segmentation network as the segmentation branch of our T-Net. Specifically, we adapted UNet3+ \cite{Huang:2020}, which was originally proposed for medical image segmentation. UNet3+ uses full-scale skip connections to capture full-scale semantics for the segmentation of organs that appear at different scales. Therefore, the temporal UNet3+ also captures full-scale details, which is extremely beneficial for tic segmentation as tics have highly variable duration.



%
\subsection{Automatic Tic Detection}
Very few studies have addressed the automatic tic 
detection
problem from video data. For instance,
\cite{Bernabei:2010} adaptively thresholded signals from triaxial accelerometers placed on the patients' trunk and achieved automatic tic detection by classifying 2-second non-overlapping windows of the recorded signals, 
\cite{Shute:2016} used deep brain stimulation (DBS) -- an invasive neuro-modulatory therapy -- to collect and classify segmented temporal windows of DBS signals using support vector machines, and
\cite{Barua:2021} used LeNet-5 to classify segmented temporal windows of wireless channel information (WCI) signals. One of the few studies that considers video data for tic 
detection
in well-trimmed videos is 
\cite{Wu:2021}, which used an unsupervised learning framework similar to SimCLR \cite{Chen:2020simclr} to extract robust video features from segmented one-second video clips and then fed these features to an LSTM for classification. However, despite the outstanding contributions of these studies, they are difficult to translate into clinical practice for two main reasons: On the one hand, non-video signals are difficult to obtain in everyday clinical settings or in patients' homes. In fact, most patients with TS would consider DBS as a last resort. On the other hand, all prior studies \cite{Bernabei:2010,Shute:2016,Barua:2021,Wu:2021} focus on detecting tics in well-trimmed signals or video clips. This limits the translational value because many tics would span between two or more windows. To the best of our knowledge, we are the first to address the problem of tic detection in untrimmed videos obtained from ubiquitous devices such as smartphones, which maximizes the translational value of our work.

\section{Methods}

Let $X = \{x_l\}_{l=1}^L$ denote an ${L}$ frame-long untrimmed video, where $x_l$ is the $l$-th frame. We denote detection annotations for $X$ as $Y = \{ y_n\}_{n=1}^N$, where $y_n = (t_n^S, t_n^E)$ is a tic instance with starting frame $t_n^S$ and ending frame $t_n^E$. For the segmentation task, we convert $Y$ to a sequence of labels $S = \{ s_l \}_{l=1}^L$, where $s_l = 1$ (tic) if $l\in[t_n^S,t_n^E]$ and $s_l = 0$ (non-tic) otherwise.

\subsection{Feature Encoding}

We adopt the two-stream style network TSN \cite{Wang:2016} to extract deep features directly from raw videos.
This network consists of two convolutional streams which process the RGB and optical flow data, respectively, followed by global average pooling. Concatenating the output of the 
two streams, we get the deep features $f_l \in \mathbb{R}^{C^{v}}$ of frame $x_l$, where $C^v = 3,072$ is the dimension of the features. 
Alternatively, we also consider interpretable features, AU intensities, extracted from the videos using
OpenFace \cite{Baltrusaitis:2018}. OpenFace extracts intensities for 17 AUs and thus we get the features $f_l \in \mathbb{R}^{C^{a}}$ of frame $x_l$, where $C^a = 17$ is the dimension of the AU intensities. 
Therefore, given an untrimmed video $X$, we can extract a feature sequence $F = \{ f_l \}_{l=1}^L \in \mathbb{R}^{C \times L}$ 
where $C \in \{ C^v, C^a \}$.

\subsection{Data Augmentation}

Since untrimmed videos are of variable temporal length and are usually too large to fit into memory for computation, we divide a video of length $L$ into $W$ video clips of fixed length $T$ using an overlapping sliding window strategy. 
Here, we use a long observation window with $T = 416$ frames to obtain $W$ feature sequences $ \{ F_w \}_{w=1}^W$, where $F_w = \{ f_t \}_{t=1}^T \in \mathbb{R}^{C \times T}$.  
We also apply random Gaussian noise 
($\mu=0, \sigma=0.01$)
with a chance of 50\% to the input features during training stage. 
\begin{figure}
\centering
\includegraphics[trim=0cm 0cm 0cm 0cm, clip,width=0.5\textwidth]{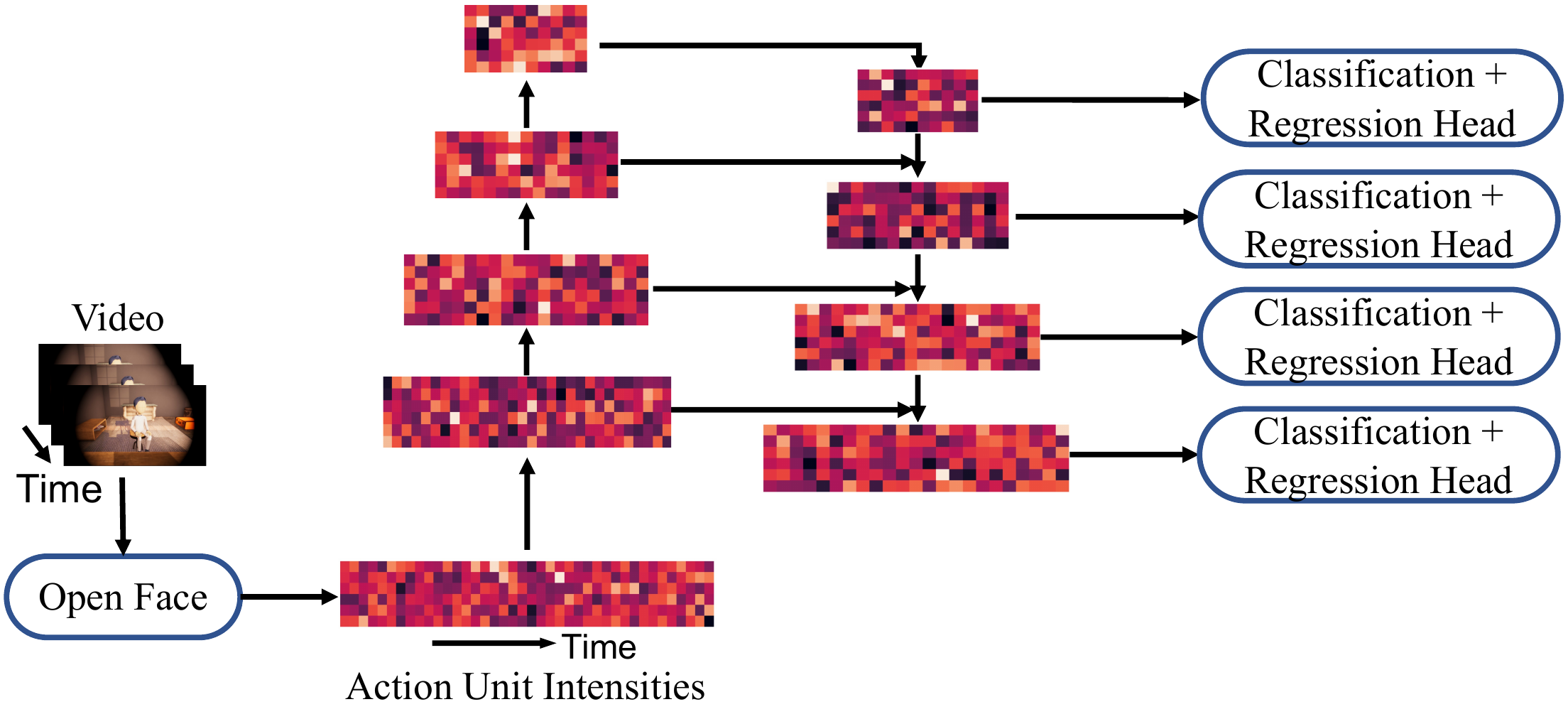}
\caption{Structure of T-RetinaNet. The system builds a 4-layer T-FPN on top of the temporal ResNet-18, followed by a two-branch convolutional head performing classification and regression.}

\label{fig:T-Retina}
\end{figure}

\subsection{T-Net}
We propose a novel one-stage temporal detector, Tic-Net (T-Net) that integrates Temporal RetinaNet (T-RetinaNet) and temporal UNet3+ (T-UNet3+) together to perform temporal tic detection. T-Net includes two branches (detection branch and segmentation branch) that share the same encoder and a score fusion head that heuristically combines the probability predictions from the two branches to generate refined probabilities for the final detections (T-Net structure is shown in Fig. \ref{fig:T-Net}). 

\subsubsection{Detection Branch}
We adapted RetinaNet \cite{Lin:2017RetinaNet} to a temporal action detector, T-RetinaNet, for detecting tics (structure shown in Fig. \ref{fig:T-Retina}). We use temporal ResNet-18 (T-ResNet-18) as the shared encoder and build a 4-stage temporal Feature Pyramid Network \cite{Lin:2017FPN} (T-FPN) on top of it as the decoder of the detection branch. In the T-FPN, The output of each ResNet residual stage is input to the corresponding stage of T-FPN to compute pyramid outputs. Then, a two-branch one-layer convolutional head is applied to the outputs of T-FPN, performing tic classification and temporal boundary window regression respectively. In summary, the detection branch $\mathcal{B}_D$ takes the features sequence $F_w$ as input and outputs detection windows with corresponding probabilities, namely 
\begin{equation}
\setlength{\abovedisplayskip}{5pt}
\setlength{\belowdisplayskip}{5pt}
\mathcal{B}_D(F_w) = D_w = \{ b_i, c_i \}_{i=1}^{N_D} \; , \ \  b_i = (\widehat{t_i^S}, \widehat{t_i^E}) \enspace,
\end{equation}
where $b_i$ denotes the $i$-th detection window with $\widehat{t_i^S}$ and $\widehat{t_i^E}$ being the starting and ending frame respectively, and $c_i$ represents the probability of $b_i$ being a tic instance.

\subsubsection{Segmentation Branch}
We adapted UNet3+ \cite{Huang:2020} to T-UNet3+ for temporal tic segmentation. T-UNet3+ utilizes the full-scale information by incorporating both the features maps from the same- and smaller-scale stages in the encoder, and the outputs from the larger-scale stages in the decoder. We build a 3-stage T-UNet3+ on top of the first three stages of T-ResNet-18 (shared encoder), namely the pre-residual stage and the first two ResNet residual stages. 
The segmentation branch $\mathcal{B}_S$ also takes the feature sequence $F_w$ as input and outputs the probability prediction for each frame at each stage, namely
\begin{equation}
\setlength{\abovedisplayskip}{5pt}
\setlength{\belowdisplayskip}{5pt}
\mathcal{B}_S(F_w) = \{ P_w^k \}_{k=1}^{K}\; , \ \ P_w^k = \{ p_j^k \}_{j=1}^T \enspace,
\end{equation}
where $K=3$ denotes the number of stages in T-UNet3+, and $p_j^k$ represents the probability of the $j$-th frame to be a tic in the $k$-th stage of the T-UNet3+ decoder.

\subsubsection{Score Fusion Head}
In the score fusion head, we combine the probability predictions from both branches heuristically to obtain refined probabilities. Specifically, we first perform an 1D RoIPooling on the output of the lowest-scale stage of the T-UNet3+ decoder, i.e. $P_w^1$, according to $\{ b_i \}_{i=1}^{N_D}$ to extract a probability vector $V_i$ for each detection window $b_i$. This probability vector is then concatenated with $c_i$, the probability prediction from the detection branch. After this, the concatenation is processed by a multilayer perceptron (MLP) to get a refined probability $\widetilde{c}_i$ to form the refined detections $\widetilde{D}_w$. Specifically, this can be expressed as
\begin{gather}
\setlength{\abovedisplayskip}{10pt}
\setlength{\belowdisplayskip}{5pt}
\widetilde{c}_i = \text{MLP} (V_i \bigoplus c_i), \ V_i=\phi(b_i, P_w^1), \ i=1,\ldots,N_D \enspace,
\\
\mathcal{B}_H(D_w, P_w^1) = \widetilde{D}_w = \{ b_i, \widetilde{c}_i \}_{i=1}^{N_D} \; , \ \ b_i = (\widehat{t_i^S}, \widehat{t_i^E}) \enspace,
\end{gather}
where $\bigoplus$ denotes concatenation and $\phi$ denotes the 1D RoIPooling operation.

\myparagraph{Anchors} Following YOLO9000 \cite{Redmon:2017}, we use k-means clustering to generate proper anchor sizes. Anchors are matched to ground-truth annotations (positives) if their efficient intersection over union (EIoU) is greater than $0.3$ or to background (negatives) if their EIoU is less than $-0.4$. EIoU \cite{Zhang:2021} is a refined version of IoU as it considers not only the IoU, but also the central points distance and lengths difference among anchors and ground-truth annotations. Specifically, we can denote an anchor as 
\begin{equation}
\setlength{\abovedisplayskip}{5pt}
\setlength{\belowdisplayskip}{5pt}
A = (t_A^S, t_A^E), \quad m_A = (t_A^S+t_A^E)/2, \quad d_A=t_A^E-t_A^S \enspace,
\end{equation}
where $t_A^S$ and $t_A^E$ represent the starting and ending frame of this anchor, respectively, $m_A$ represents the central point of this anchor, and $d_A$ represents the length. Similarly, we can also denote a ground-truth annotation $G = (t_G^S, t_G^E)$ with central point $m_G$ and length $d_G$, and the smallest enclosing window $C = (t_C^S, t_C^E)$ where $t_C^S = \min\{ t_A^S, t_G^S \}$ and $t_C^E = \max\{ t_A^E, t_G^E \}$, with central point $m_C$ and length $d_C$. 
Finally, the EIoU between the anchor and the annotation can be expressed as
\begin{equation}
\setlength{\abovedisplayskip}{5pt}
\setlength{\belowdisplayskip}{5pt}
\text{EIoU} = \text{IoU} - \frac{(m_A-m_G)^2} { d_C^2} - \frac{(d_A-d_G)^2} {d_C^2} \enspace.
\end{equation}

\myparagraph{Training} For the detection branch, we first perform online hard negative mining as used in \cite{Liu:2016} with a positive-to-negative ratio of $1:3$ to address the imbalance problem. Then, we apply binary cross-entropy (bce) loss for classification to all the positives and negatives and $\text{SmoothL}_1$ ($\text{sl}$) loss for regression to all the positives. The implementation of these losses is the same as in \cite{Liu:2016}. Additionally, we apply EIoU loss \cite{Zhang:2021}, namely $1-\text{EIoU}$, for regression as well, which encourages the detections to be as similar as possible to the ground-truth annotations. The total loss for the detection branch can be expressed as the weighted sum of the classification loss and the regression losses, namely
\begin{equation}
\setlength{\abovedisplayskip}{5pt}
\setlength{\belowdisplayskip}{5pt}
L_w^D = \frac{1}{N_p} \sum\nolimits_i L_{w,{\text{bce}}}(\widetilde{c}_i, \widetilde{y}_i) + \gamma (L_{w,\text{sl}} + \lambda L_{w,\text{EIoU}}) (b_i, y_i) \; ,
\end{equation}
where $\widetilde{y}_i = 1$ if the anchor for generating $b_i$ was matched to a ground-truth (positive) otherwise $\widetilde{y}_i = 0$, and $y_i = (t_i^S, t_i^E)$ denotes the ground-truth annotation. $N_p$ is the number of positives. For the hyperparameters, we set $\gamma=2$ and $\lambda=1.5$.

For the segmentation branch, we adopt focal loss (fl) \cite{Lin:2017RetinaNet} and Dice loss and use deep supervisions
on the predicted probability sequences from all stages in T-UNet3+, namely $\{ P_w^k \}_{k=1}^K$, following \cite{Huang:2020}. 
The loss for the segmentation branch can be expressed as 
\begin{equation}
\setlength{\abovedisplayskip}{5pt}
\setlength{\belowdisplayskip}{5pt}
L_{w}^S = \frac{1}{T}(L_{w,\text{fl}}  + L_{w, \text{Dice}})  (\{ P_w^k \}_{k=1}^K, \{ s_j \}_{j=1}^T) \enspace.
\end{equation}

Finally, the total loss for training T-Net can be expressed as the weighted sum of the detection branch loss, the segmentation branch loss, and $L_2$ regularization term, 
\begin{equation}
\setlength{\abovedisplayskip}{5pt}
\setlength{\belowdisplayskip}{5pt}
L_w = L_{w}^D + \alpha L_{w}^S + \beta L_2(\Theta) \enspace,
\end{equation} 
where we set $\alpha=3$ and $\beta=0.001$.

To enlarge the model capacity, we also allow independent encoders for detection and segmentation branch separately, which is referred to as T-Net (independent) while the shared-encoder version is T-Net (shared).

\myparagraph{Inference} Detections from all pyramidal levels are merged with a probability threshold of 0.2 and NMS using EIoU with a threshold of 0.2 is applied to yield final detections.


\section{Experiments}

\subsection{Data Acquisition and Annotation Protocol}
Facial motor tics of one 12-year-old male diagnosed with TS were analyzed in this paper. 
We recorded 12 15-minute videos across four visits for this participant. 
In each visit, he performed three activities. In Activity 1, the participant sat in a chair and did not engage in any intentional movement. In Activity 2, the participant was engaged in a movement-based activity (e.g., completing schoolwork). In Activity 3, the participant was engaged in conversation with an experimenter about tics.
All videos were recorded using a smart phone with 30 FPS. Thereafter, temporal boundary windows were annotated using open-source software VIA3 \cite{Dutta:2019}. Each video was first labelled by two trained annotators. Then the two sets of annotations were checked by a senior researcher, who resolved any inconsistency.

\subsection{Cross-validation Strategies}
In total, 2631 tic instances were annotated from the entire 12 videos. Mean and median duration of annotations were 41.15 and 32 frames, respectively, and 99.96\% of the annotations were shorter than 416 frames.
Experiments were performed following two cross-validation strategies: (i) \textbf{Leave One Session Out (LOSO):} one out of four sessions (three out of twelve videos) were held for testing, and (ii) \textbf{Leave One Video Out (LOVO):} one out of twelve videos was held for testing.

\definecolor{Gray}{gray}{0.85}

\begin{table}
\centering
\resizebox{\columnwidth}{!}{%
\begin{tabular}{c c c c}
\toprule
\textbf{Features} & \textbf{Methods} & \multicolumn{2}{c}{\textbf{AP@$0.5$}}\\
& &LOSO & LOVO \\
\midrule
Deep Features & BMN \cite{Lin:2019} 	&	\textbf{40.32}	&	\textbf{45.87}	\\
Deep Features & DaoTAD \cite{Wang:2021daotad} 	&	35.52	&	38.29	\\
Deep Features & TCANet \cite{Qing:2021tcanet} 	&	39.40	&	40.30	\\


\midrule

AU intensities & BMN \cite{Lin:2019} 	&	33.09	&	38.26	\\
AU intensities & DaoTAD \cite{Wang:2021daotad}	&	10.19	&	3.29	\\
AU intensities & TCANet \cite{Qing:2021tcanet}	&	16.42	&	30.02	\\

\rowcolor{Gray}
AU intensities & T-Net (detection branch only) 	&	16.33	&	25.08	\\
\rowcolor{Gray}
AU intensities & T-Net (shared) 	&	35.92	&	41.84	\\
\rowcolor{Gray}
AU intensities & T-Net (independent) 	&	\textbf{37.30}	&	\textbf{43.73}	\\
\bottomrule
\end{tabular}
}
\caption{Detection results: AP at IoU threshold of 0.5 and probability threshold of 0.2. The metrics are averaged across splits for each cross-validation strategy.  \label{detection_results}}
\vspace{-1em}
\end{table}

\subsection{Detection Results}
Table \ref{detection_results} presents the metric, average precision (AP), for both cross-validation strategies. The metric is calculated at the IoU threshold of 0.5 and probability threshold of 0.2, i.e. only detections with predicted probability (of this detection being a tic) greater than 0.2 will be considered and the tic is considered detected if the IoU between the detection and the ground-truth annotation is larger than 0.5. We evaluate the baseline systems based on BMN \cite{Lin:2019}, DaoTAD \cite{Wang:2021daotad} and TCANet \cite{Qing:2021tcanet} that use deep features extracted from raw videos and they achieve the AP of 45.87, 38.29 and 40.30 respectively in LOVO. This shows that we successfully established workable systems for temporal tic detection in untrimmed videos. 
Moreover, LOVO results show significant improvements as compared to LOSO results, which indicates getting more data could be beneficial to promote detection performance.



Besides, when comparing T-Net with baseline methods, T-Net that operates on AU intensities outperforms all three baseline methods using AU intensities as input features. For comparison with the baseline methods using deep features, T-Net achieves comparable performance (AP of 43.73) to BMN (AP of 45.87) and better performance than DaoTAD (AP of 38.29) and TCANet (AP of 40.30) in LOVO. Therefore, these results show the effectiveness of our proposed T-Net which can improve the interpretability 
through replacing deep features with AU intensities without significantly affecting the performance.

Moreover, the grey-shaded rows in Table \ref{detection_results} present an ablation analysis where we compare the usefulness of adding the segmentation branch and enlarging model capacity in T-Net. Having the additional segmentation branch in T-Net (shared) brings around 19\% and 16\% increase in LOSO and LOVO respectively, compared to using the detection branch only. We also investigated the results and found the refinement of probability significantly reduces many high-probability false positives and therefore improves the performance. This shows the refinement of probability is crucial and effective for T-Net. Besides, enlarging the model capacity by allowing independent encoders for detection and segmentation branch, namely T-Net (independent) brings around 2\% increase compared to T-Net (shared), which implies the detection branch and the segmentation branch require different type of features and by having different encoders, branch-specific instead of general features can be extracted.

\subsection{Interpretability of AU intensities}
In clinical practice, clinicians identify motor tics by evaluating the forcefulness, abruptness, and location of patient movements. This includes a combination of the intensity of muscle activations, identity and combination of the muscle involved, and the temporal relationships between these activations. As AUs represent the basic movements of facial muscles or muscle groups, AUs capture the spatial aspect of the features clinicians use to identify tics. Since AUs represent a meaningful analog to how tics are identified by clinicians, models using AU intensities as features would in turn be more interpretable by humans, unlike visual features extracted by deep neural networks. 

\begin{figure}
\centering
\includegraphics[trim=0cm 0cm 0cm 0cm, clip,width=\columnwidth]{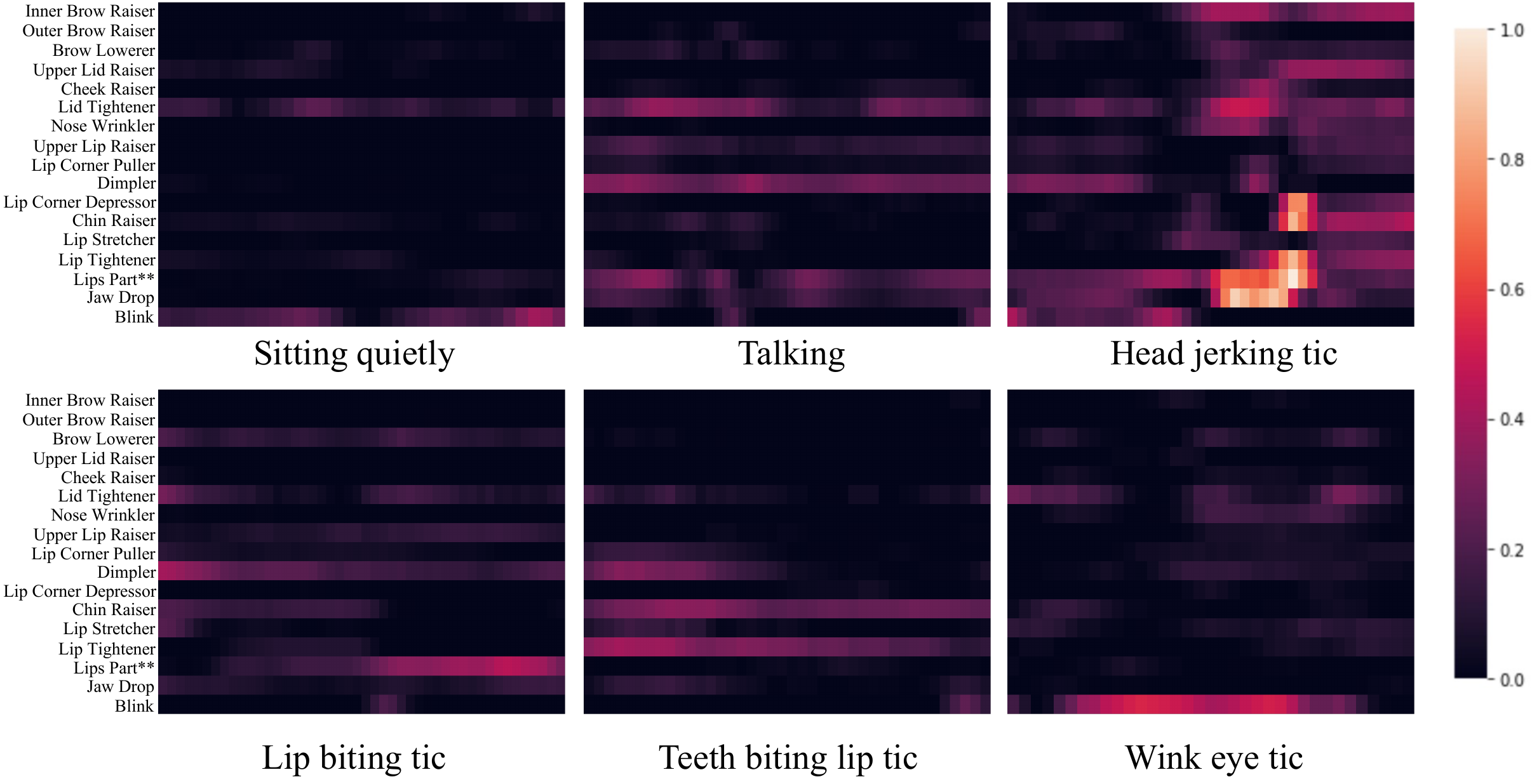}
\caption{Normalized AU intensities heatmap for example actions.}
\label{fig:AU_interpretability}
\end{figure}

Fig. \ref{fig:AU_interpretability} illustrates the interpretability of AU intensities. Here, heatmaps show the normalized AU intensities of six actions. The y-axis lists the 17 AUs extracted using OpenFace (e.g., ``inner brow raiser", ``upper lid raiser", ``chin raiser"). The x-axis denotes time frames and the color represents the normalized AU intensities (brighter colors indicate stronger intensity). When the participant sat quietly (non-tic behavior, top row left), ``Blink" was the only AU moderately activated. When participant talked (non-tic behavior, top row middle), AUs related to lips and mouth were activated whereas the other AUs showed low intensity. The other four maps correspond to specific tics of this participant. During his ``Head jerking tic", the participant would suddenly twist his head, and the heatmap shows nearly all AUs were activated, with the highest intensities observed for AUs involving the jaw, mouth, and lips. Likewise, AUs involving the lips were strongly activated during the participant’s ``Lip biting tic" (upper lip pressing down on the lower lip) and ``Teeth biting lip tic" (teeth biting the lower lip). Finally, during the ``Wink eye tic" (closing one eye forcefully), the AU ``(Eye)Lid Tightener" was activated and the AU ``Blink" exhibited higher intensity than that during the “sitting quietly” activity, in which the same AU was only moderately activated.

\begin{figure}
\centering
\includegraphics[trim=0cm 0cm 0cm 0cm, clip,width=\columnwidth]{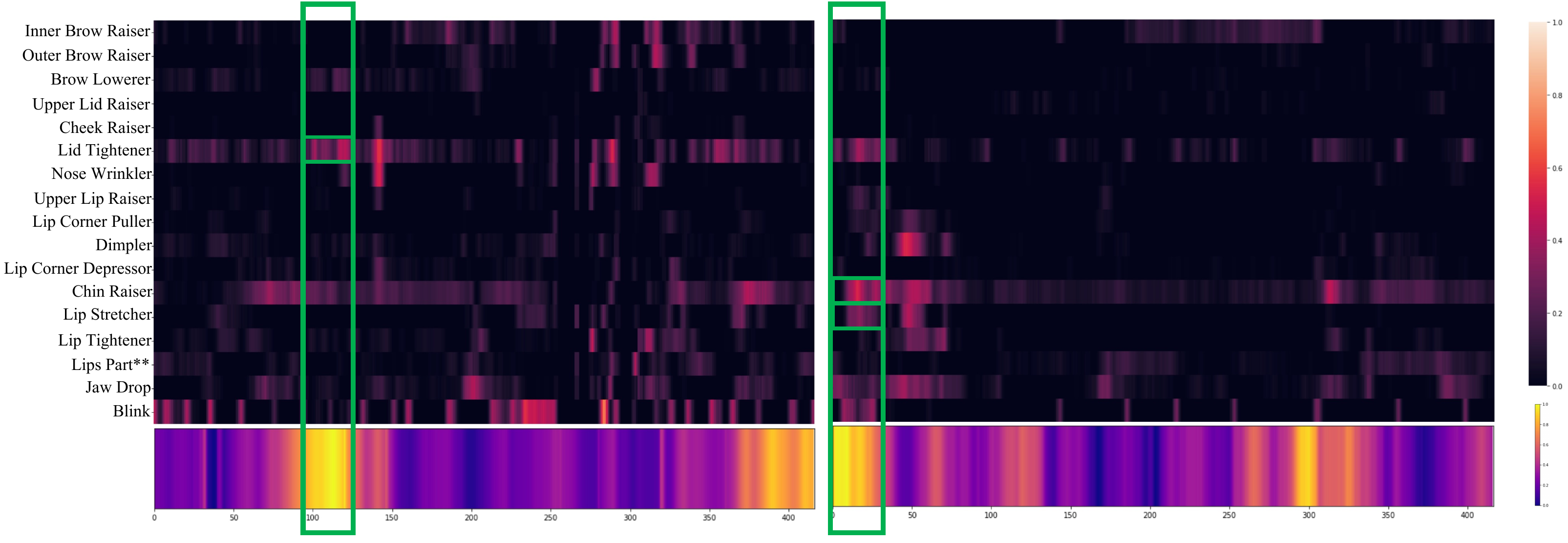}
\caption{\textbf{Top:} normalized AU intensities heatmap. \textbf{Bottom:} 1D-Grad-CAM visualization. Brighter means more informative.}
\label{fig:AU_gradcam}
\end{figure}

Fig. \ref{fig:AU_gradcam} illustrates two examples showing how the T-Net captures the patterns of AU intensities using 1D-Grad-CAM. Grad-CAM was designed to visually explain classifiers' decisions, which highlights spatially important regions in 2D images. It uses the gradients to weigh feature maps which are then combined to get the class activation map (CAM). In this study, we adapted it to 1D domain (1D-Grad-CAM) to map important time frames to the detector's classification decision (i.e. whether a temporal window contains a tic). The top row in Fig. \ref{fig:AU_gradcam} shows heatmap of normalized AU intensities for an observation window 
where AUs are listed in the y-axis and the x-axis represents time frames. 
The bottom row shows the visualization of 1D-Grad-CAM results. In the left example, we have a detection from 85.40 to 126.86 frame which contained a ``Wink eye tic". From the figure, the CAM is focused on the area around the detection which highlights the strongly activated AU ``(Eye)Lid Tightener" marked by a green box on the heatmap. In the right example, a detection starting from 0.00 to 26.34 frame contained a ``Lip biting tic" where CAM also shows focusing on the area and lips and mouth related AUs are activated (marked by green boxes). Thus, the AU intensities themselves are interpretable and our T-Net detector captures the interpretable patterns of the AU intensities which subsequently 
brings more clinical interpretability to our T-Net.

\section{Conclusion}
In this work, we were the first to develop a system, T-Net that combined segmentation probabilities to refine detection probabilities, for automatic facial tic detection in untrimmed videos. 
T-Net helped lift the barriers to the behavioral therapy by promoting accessibility and providing accurate tic detections.
T-Net was easy-to-implement because the videos used were recorded by a ubiquitous device, a smartphone. T-Net was also shown to be accurate as it achieved comparable performance to baseline methods which use deep features. Finally, T-Net was interpretable as it used interpretable AU intensities as input features and captured the interpretable patterns of AU intensities.



\section*{Acknowledgment}
This work has been supported by Discovery Challenge Award, Johns Hopkins Discovery Fund Program.


\bibliographystyle{./IEEEtran}
\bibliography{./Tic_ref}

\end{document}